\documentclass[runningheads]{llncs}

\usepackage{eccv}

\usepackage[table]{xcolor}

\usepackage{eccvabbrv}

\usepackage{graphicx}
\usepackage{booktabs, multirow}

\usepackage[accsupp]{axessibility}  

\usepackage{hyperref}

\usepackage{orcidlink}

\begin{document}

\title{Do Vision Foundation Models Enhance Domain Generalization in Medical Image Segmentation?}

\titlerunning{Do Vision FMs Enhance Domain Generalization?}

\author{Kerem Cekmeceli\inst{1}$^\dag$\and
Meva Himmetoglu\inst{1} \and
Guney I. Tombak\inst{1} \and
Anna Susmelj\inst{1, 2, \star} \and
Ertunc Erdil\inst{1, \star} \and
Ender Konukoglu \inst{1, 3, \star}
}

\authorrunning{K. Cekmeceli et al.}

\institute{Computer Vision Lab., ETH Zurich \and
ETH AI Center \and
The LOOP Zürich – Medical Research Center, Zürich, Switzerland \\
\email{$^\dag$kcekmeceli@student.ethz.ch}
}

\renewcommand{\thefootnote}{\fnsymbol{footnote}} 

\maketitle

\footnotetext[1]{Joint supervision}

\begin{abstract}
Neural networks achieve state-of-the-art performance in many supervised learning tasks when the training data distribution matches the test data distribution. 
However, their performance drops significantly under domain (covariate) shift, a prevalent issue in medical image segmentation due to varying acquisition settings across different scanner models and protocols. 
Recently, foundational models (FMs) trained on large datasets have gained attention for their ability to be adapted for downstream tasks and achieve state-of-the-art performance with excellent generalization capabilities on natural images.
However, their effectiveness in medical image segmentation remains underexplored.
In this paper, we investigate the domain generalization performance of various FMs, including DinoV2, SAM, MedSAM, and MAE, when fine-tuned using various parameter-efficient fine-tuning (PEFT) techniques such as Ladder and Rein (+LoRA) and decoder heads. 
We introduce a novel decode head architecture, HQHSAM, which simply integrates elements from two state-of-the-art decoder heads, HSAM and HQSAM, to enhance segmentation performance. 
Our extensive experiments on multiple datasets, encompassing various anatomies and modalities, reveal that FMs, particularly with the HQHSAM decode head, improve domain generalization for medical image segmentation. 
Moreover, we found that the effectiveness of PEFT techniques varies across different FMs.
These findings underscore the potential of FMs to enhance the domain generalization performance of neural networks in medical image segmentation across diverse clinical settings, providing a solid foundation for future research. Code and models are available for research purposes at \url{https://github.com/kerem-cekmeceli/Foundation-Models-for-Medical-Imagery}.
\keywords{Foundational models \and Medical image segmentation \and Domain generalization}
\end{abstract}

\section{Introduction}
\label{sec:intro}
Despite the impressive performance of neural networks across various domains, their ability to generalize to data distributions different from those on which they were trained remains limited \cite{learningdataset}. This issue, known as domain or distribution shift, is particularly prevalent in medical imaging, where data distribution can vary significantly between different imaging modalities or protocols. To address this challenge, significant efforts have been made to enhance neural network performance under distribution shift using techniques such as transfer learning \cite{van2014transfer, tajbakhsh2016convolutional, karani2018lifelong}, unsupervised domain adaptation \cite{kamnitsas2017unsupervised, ouyang2019data, dou2019pnp}, test-time adaptation \cite{karani2021test, bateson2022test, valanarasu2024fly}, and domain generalization \cite{huo2018synseg, dou2019domain, zhang2020generalizable}.

FMs have garnered significant attention due to their remarkable generalization capabilities in Natural Language Processing (NLP) \cite{devlin2018bert, lewis2019bart, brown2020language, radford2018improving, radford2019language, bommasani2021opportunities}. Their success is largely attributed to their ability to learn from vast datasets and the extensive number of parameters they possess. This success in NLP is now inspiring other fields, including computer vision \cite{dosovitskiy2020image, radford2021learning, kirillov2023segment, oquab2023dinov2, ke2024segment} and medical imaging \cite{ma2024segment, butoi2023universeg, wang2022medclip}, with the goal of achieving a comparable transformative impact. FMs exhibit exceptional zero-shot performance and provide representations that achieve state-of-the-art results with superior generalization capabilities when fine-tuned for downstream tasks. 
However, despite their promising performance, a comprehensive comparison of their effectiveness on out-of-distribution tasks in medical image segmentation remains lacking.

Adapting FMs for downstream tasks in medical image segmentation often involves fine-tuning on small or mid-sized labeled datasets due to the limited availability of large labeled datasets \cite{chaitanya2020contrastive}. Fine-tuning all parameters on such datasets can lead to overfitting and reduced generalization \cite{wei2024stronger}, and is also prohibitively expensive. Parameter-efficient fine-tuning (PEFT) techniques address this issue by updating only a small number of parameters while keeping the rest frozen \cite{hu2021lora, chai2023ladder, wei2024stronger}. The effectiveness of various PEFT methods in combination with different FMs for domain generalization in medical image segmentation remains an open question.

Decoding representations from FMs into segmentation masks requires an effective decoder architecture. Although various decoding heads have been proposed \cite{ronneberger2015u, kirillov2023segment, ke2024segment}, their effectiveness for medical image segmentation when fine-tuning different FMs is not well understood.

In light of these observations, this paper makes the following contributions:
\begin{itemize}
    \item We investigate the domain generalization performance of various FMs (DinoV2 \cite{oquab2023dinov2}, SAM \cite{kirillov2023segment}, MedSAM \cite{ma2024segment}, and MAE \cite{he2022masked}), when fine-tuned using various PEFT techniques (Ladder \cite{chai2023ladder}, Rein \cite{wei2024stronger}, and Rein-Lora \cite{wei2024stronger}) and decoder heads (Linear \cite{oquab2023dinov2}, Resnet-like \cite{he2016deep}, Unet-like \cite{ronneberger2015u}, Dual-attention (DA) head \cite{fu2019dual}, Segformer \cite{xie2021segformer}, SAM mask decoder \cite{kirillov2023segment}, HSAM \cite{cheng2024unleashing}, and HQSAM \cite{ke2024segment}). 

    \item We introduce a novel decode head architecture, HQHSAM, which simply integrates elements from two state-of-the-art decoder heads, HSAM and HQSAM, and show that it enhances the domain generalization performance.

     \item We present extensive experiments on multiple datasets encompassing various anatomies (healthy and abnormal brain, prostate, and lumbar spine) and modalities (T1w, T2w, FLAIR, and CT). 
\end{itemize}

\section{Related Work}
\label{sec:related}

\subsection{Foundational Models}
\label{sec:related_fms}
FMs can be broadly classified into two types: feature encoding models trained on pretext tasks, such as Dino \cite{caron2021emerging}, DinoV2 \cite{oquab2023dinov2}, Masked Autoencoders (MAE) \cite{he2022masked}, CLIP \cite{radford2021learning}, BLIP \cite{li2022blip}, and models trained for specific tasks such as SAM \cite{kirillov2023segment}, and SEEM \cite{zou2024segment} for image segmentation.

Despite the significant advancements in FMs within NLP and computer vision, their impact on medical imaging has been limited. One of the primary reasons for this is the difficulty in accessing and aggregating large-scale datasets comparable to those used in other domains. Medical imaging data are often fragmented across different institutions and subject to stringent privacy regulations, making it challenging to amass the extensive datasets required for training large FMs. Many efforts to exploit FMs for medical image segmentation have involved applying the out-of-the-box SAM model to various datasets \cite{deng2023segment, hu2023sam, he2023accuracy, roy2023sam, zhou2023can, mohapatra2023sam}. Comprehensive assessments of SAM on a diverse array of medical images indicate that SAM achieves satisfactory segmentation on structures with distinct boundaries but struggles with boundaries that are weak and have low contrast, which is prevalent in medical images \cite{mazurowski2023segment}. MedSAM \cite{ma2024segment} was proposed as a specialized FM for medical image segmentation, achieved by fine-tuning the SAM model on a large-scale dataset containing more than one million medical image-mask pairs, which was curated from publicly available datasets.

\subsection{Parameter-efficient fine-tuning (PEFT) methods}
\label{sec:related_peft}
Low-Rank Adaptation (LoRA) \cite{hu2021lora}, initially proposed for NLP, injects trainable rank decomposition matrices into each layer of the Transformer architecture, updating only these parameters during fine-tuning while keeping the rest frozen. Ladder fine-tuning \cite{chai2023ladder} combines the representation of the SAM model's encoder with that obtained from a convolutional neural network (CNN) encoder. The combined representations are decoded into a segmentation, and during fine-tuning, only the CNN encoder and decoder are updated while the SAM encoder remains frozen. Rein \cite{wei2024stronger}, a PEFT method for domain generalization in semantic segmentation, consists of a collection of learnable token sequences that interact with distinct instances to facilitate instance-level feature refinement. These learnable token sequences can also be represented as the multiplication of two low-rank matrices to achieve further parameter efficiency, referred to as Rein-Lora \cite{wei2024stronger}.

\section{Method}
\label{sec:method}

\subsection{Datasets}
\label{sec:datasets}
\noindent \textbf{Healthy Brain MRI:} We used two publicly available datasets: the Human Connectome Project (HCP) \cite{van2013wu} and the Autism Brain Imaging Data Exchange (ABIDE) \cite{di2014autism}. The HCP dataset includes both T1-weighted (T1w) and T2-weighted (T2w) images for each subject, while the ABIDE dataset contains T1w images from various centers. For our study, we selected the subsets from Caltech (ABIDE-C) and Stanford (ABIDE-S). Since these datasets lack manual segmentations, we generated labels for 15 subcortical structures using the FreeSurfer tool \cite{fischl2012freesurfer}. The labels include background, cerebellum gray matter, cerebellum white matter, cerebral gray matter, cerebral white matter, thalamus, hippocampus, amygdala, ventricles, caudate, putamen, pallidum, ventral DC, CSF, and brain stem.

\noindent \textbf{Prostate:} We utilized datasets from National Cancer Institute \cite{bloch10nci} and an in-house dataset. Each subject in the datasets includes expert annotations for three labels: background, central gland (CG), and peripheral zone (PZ). 

\noindent \textbf{Lumbar spine:} We used two publicly available datasets namely VerSe \cite{sekuboyina2021verse} and MrSegV \cite{al2019boundary}. Both datasets contain labels for six classes. 

\noindent \textbf{Brain Tumor:} For brain tumor segmentation, we utilize a subset of the BraTS dataset \cite{menze2014multimodal}, which includes T1-weighted (T1w) and FLAIR images for each subject. The available labels cover necrotic core, non-enhancing core, enhancing core, and edema, and were reduced to binary labeling of lesion and background for our experiments. 
The dataset details are summarized in Table \ref{tab:dataset_summary}. Note that some of the datasets were used during the training of the MedSAM model. Therefore, they are excluded from certain experiments involving the MedSAM backbone.
\begin{table}[h]
\centering
\caption{Dataset details. * indicates that the dataset is used for MedSAM\cite{ma2024segment} training}
\footnotesize
\begin{tabular}{c|c|c|c|c|c}
    \textbf{Anatomy} & \textbf{Dataset} & \textbf{Modality} & \textbf{$N_{train}$} & \textbf{$N_{val}$} & \textbf{$N_{test}$} \\ \hline
    \multirow{4}{*}{Healty Brain} & HCP & T1w MRI & 20 & 5 & 20 \\ 
                           & HCP & T2w MRI & 20 & 5 & 20 \\ 
                           & ABIDE-C & T1w MRI & 10 & 5 & 20 \\ 
                           & ABIDE-S & T1w MRI & 10 & 5 & 20 \\ \hline
    \multirow{2}{*}{Prostate} & NCI* & \multirow{2}{*}{T2w MRI} & 10 & 5 & 15 \\ 
                              & In-house & & 28 & 20 & 20 \\ \hline
    \multirow{2}{*}{Lumbar Spine} & VerSe & CT & 116 & 1 & 15 \\ 
                                  & MRSegV & T1w MRI & 162 & 20 & 20 \\ \hline
    \multirow{2}{*}{Brain Tumor} & BraTS* & T1w MRI & 198 & 30 & 57 \\ 
                                 & BraTS* & FLAIR MRI& 198 & 30 & 57 \\ 
\end{tabular}
\label{tab:dataset_summary}
\end{table}

\subsection{Encoders}
\label{sec:encoders}
In our experiments, we used DinoV2 \cite{oquab2023dinov2}, SAM \cite{kirillov2023segment}, MedSAM \cite{ma2024segment}, and MAE \cite{he2022masked} encoders as feature extractors. To ensure that the input size of the encoders is independent from the input size, we implemented bilinear interpolation for the positional encoding across all FMs, similar to the approach used in \cite{oquab2023dinov2}. This enhancement is necessary to adjust the expected image size of FMs to medical images.

\subsection{Decoders}
\label{sec:decoders}

\noindent \textbf{Linear decoder} consists of a simple 2D Batch Normalization and $1\times1$ convolution used to provide a simple and efficient way to map the feature representations from the encoder network to the segmentation mask.

\noindent \textbf{ResNet decoder} consists of residual blocks motivated by the ResNet architecture \cite{he2016deep}. In each block, the input is upscaled using transpose convolution which is then passed to residual blocks consisting of $3\times3$ convolutions, ReLU non-linearity, and Batch Normalization, repeated $M$-times. The final feature map is passed through $1\times1$ convolutions to obtain the segmentation map.

\noindent \textbf{UNet decoder} is adapted for transformer-based encoders following the original approach described in \cite{ronneberger2015u}. The encoder's output features are processed through a series of residual blocks and then upscaled. These upscaled features are concatenated with features from the previous layer of the FM, mimicking the skip connections in \cite{ronneberger2015u}, and fed into another series of residual blocks. This process continues until the original image resolution is achieved. Finally, a $1\times1$ convolution, similar to that in ResNet, is applied to generate the segmentation map.

\noindent \textbf{Dual-attention (DA) decoder} is proposed to adaptively integrate local features with their global dependencies \cite{fu2019dual}. We used the MMSEG \cite{mmseg2020} implementation in our experiments. 

\noindent \textbf{SegFormer decoder} combines hierarchical feature representations from multiple network stages with lightweight multi-layer perceptron (MLP) decoders, enabling efficient and accurate segmentation \cite{xie2021segformer}. This design allows the model to capture both fine details and global context, improving overall segmentation performance. We used the MMSEG \cite{mmseg2020} implementation in our experiments. 

\noindent \textbf{SAM Mask Decoder} is a powerful yet efficient transformer-based decoder design introduced in \cite{kirillov2023segment} and has been adapted to be compatible with any backbone in our implementation. The decoder takes input from a backbone and a prompt encoder. We operated this decoder prompt-free by using the default prompt. In our experiments, we both tried training the prompt encoder (SAMMD-PE) and keeping it frozen (SAMMD-FPE). We initialized the prompt encoder and the mask decoder with the pretrained weights from either SAM or MedSAM depending on the encoder. 

\noindent \textbf{HQSAM Decoder} enhances the original SAM decoder by introducing a high-quality (HQ) output token and a new mask prediction layer \cite{ke2024segment}. This setup refines mask predictions by integrating features from both the first and last stages of the model, thereby enhancing detail and accuracy in segmentation. 

\noindent \textbf{HSAM Decoder} incorporates a two-stage hierarchical decoding process that enhances the segmentation quality \cite{cheng2024unleashing}. It refines the initial mask predictions with mask-guided self-attention and learnable mask cross-attention. 

\begin{figure}
    \centering
    \includegraphics[width=\textwidth]{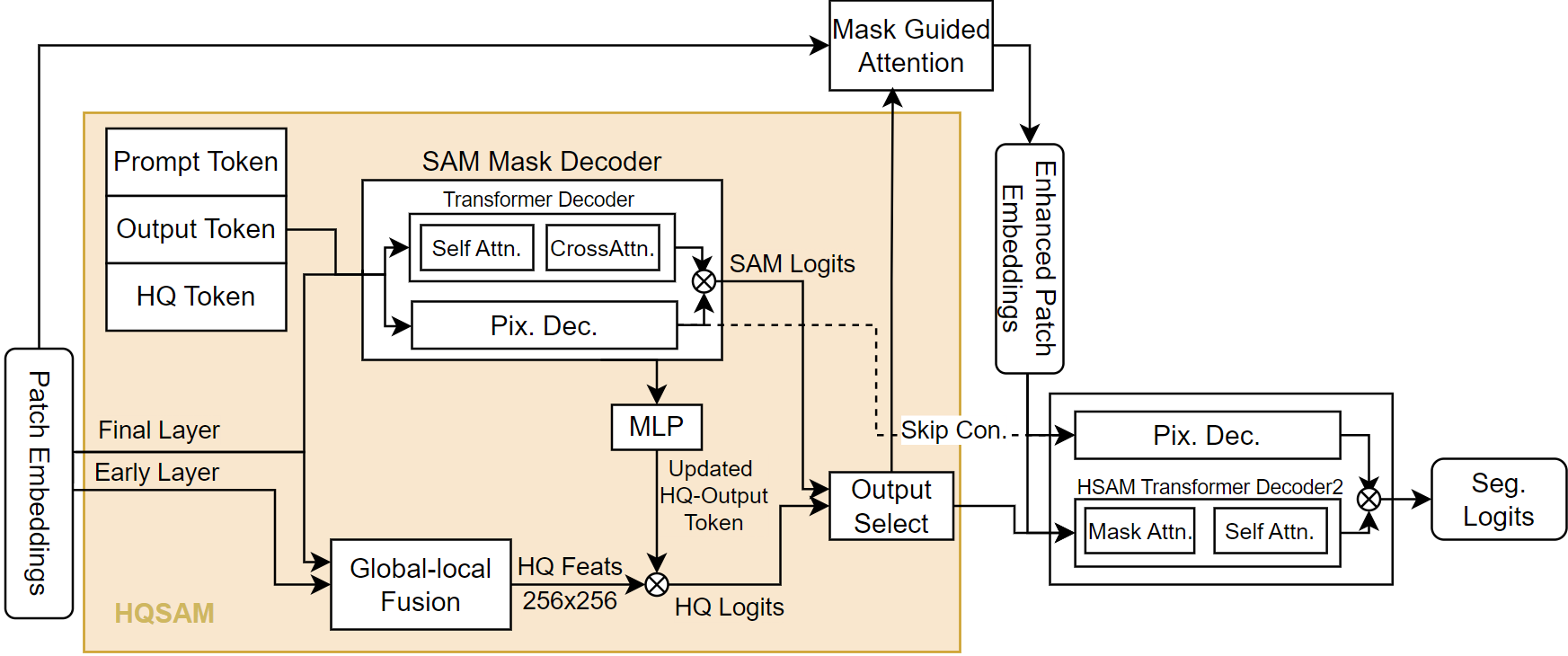}
    \caption{The architecture of the proposed HQHSAM Decoder which integrates elements from HSAM and HQSAM.}
    \label{fig:hqhsam}
\end{figure}

\noindent \textbf{HQHSAM Decoder} is our proposed decoder head, which combines elements from HQSAM and HSAM as illustrated in Figure \ref{fig:hqhsam}. HQHSAM integrates the HQ token and leverages features from various depths of the backbone within the dual-stage decoding architecture of the HSAM projection head. The output from the HQSAM block is fed into the secondary decoder block of HSAM.

\subsection{Implementation details}
\noindent\textbf{Preprocessing:} Initially, we standardized the voxel sizes of the healthy brain, lumbar spine, and prostate datasets to $0.7 \times 0.625 \times 0.7 \mathrm{mm}^3$. For the brain tumor datasets, we first employ affine registration to MNI space \cite{mni} using ANTsPy \cite{ants}, then we interpolated the image sizes of each slice to $256 \times 256$.

We performed N4 bias field correction \cite{tustison2010n4itk} on all MRI datasets. Then, the volumes were normalized using the 1st ($x_p^1$) and 99th ($x_p^{99}$) percentiles with the formula $x_{normalized} = (x - x_p^1) / (x_p^{99} - x_p^1)$, except for the BraTS datasets, which were subjected to Z-normalization using $x_{znorm} = (x - \mu) / \sigma$ followed by min-max normalization $x_{normalized} = (x_{znorm} - \min(x_{znorm})) /(\max(x_{znorm}) - \min(x_{znorm}))$. Skull stripping was applied to both the healthy brain and brain tumor datasets.

\noindent\textbf{Data augmentation:} During training, we applied extensive data augmentation to enhance the models' robustness and generalization performance. Specifically, we applied brightness, contrast, saturation, and hue adjustments with a probability of 0.5, elastic transformation with a probability of 0.25, and rotation, translation, and scaling with a probability of 0.25.

\noindent\textbf{Hyperparameters:} We trained the models using the AdamW optimizer \cite{loshchilov2017decoupled} with a weight decay of 1e-5, keeping all other parameters at their default settings. A linearly decaying learning rate was employed, with a linear warm-up for the initial 5\% of the total epochs. The initial learning rate was selected based on the backbone and dataset to ensure stability: either 1e-5, 2e-5, or 5e-5. By default, a learning rate of 5e-5 was used, except for the Dino backbone applied to the HCP and VerSe datasets, where a learning rate of 1e-5 was set. For the other datasets using the Dino backbone, a learning rate of 2e-5 was used. Additionally, when Reins fine-tuning was applied to any backbone, a learning rate of 2e-5 was utilized.

The BraTS and VerSe datasets are trained for 80 and 100 epochs, respectively, while the rest of the models are trained for 120 epochs. Cross-entropy loss is used for all experiments. Training is performed using a single NVIDIA A6000 GPU. Our implementation is based on PyTorch-Lightning.

\noindent\textbf{Evaluation metric:} We used the Dice score \cite{dice1945measures} in all experiments to measure the overlap between the predicted and ground truth segmentations. The Dice score ranges from 0 to 1, with higher values indicating better performance.

\section{Experiments}

\subsection{Decoder selection}
In this experiment, we assessed both in-domain (ID) and domain generalization (DG) performance of all decoders (see Section \ref{sec:decoders}) paired with foundational model (FM) encoders (see Section \ref{sec:encoders}). To streamline the number of runs, we did not investigate any parameter-efficient fine-tuning (PEFT) methods during the decoder selection experiments. Instead, we kept the FM parameters frozen and only updated the decoder parameters during training. Additionally, we used the base versions of all FMs. The experiments were conducted on datasets encompassing healthy brain, prostate, and lumbar spine. Specifically, we selected the ABIDE-C, In-house, and MRSegV datasets as the source domain datasets, and evaluated the domain generalization performance on the remaining datasets of the corresponding anatomy.

In Table \ref{tab:results_decoder}, we present a summary of the quantitative results. We averaged the Dice scores across all backbones and datasets for each decoder head to provide a clear comparison between decoders. The results indicate that decoders utilizing SAM's prompt encoder and mask decoder architecture consistently outperform others across all evaluated categories. Notably, HSAM and HQSAM show further performance enhancements. Our proposed decoder, HQHSAM, offers slight additional improvements. Based on this evaluation, we have decided to use HQHSAM as the decoder for the FMs in the remaining experiments. The more detailed results for each backbone, dataset, and decoder head can be found in the supplementary material.
\begin{table}[h]
    \centering
    \caption{Comparisons of different decoders averaged over all FMs and datasets. (*) \textit{Shown for brain datasets using a patch size of 16}. The best three results in each category (ID and DG) are highlighted by bold.}
    \label{tab:results_decoder}
    \begin{tabular}{c|c|c|c}
        Decoder & \#Trainable Params$^*$ & \cellcolor{gray!20} ID & \cellcolor{gray!70}DG \\ \hline
        HQHSAM  & 17.32M    & \cellcolor{gray!20}\textbf{84.98}    & \cellcolor{gray!70}\textbf{52.63} \\ 
        HQSAM   & 7.61M     & \cellcolor{gray!20}\textbf{84.88}    & \cellcolor{gray!70}\textbf{52.18} \\ 
        HSAM    & 15.98M    & \cellcolor{gray!20}\textbf{83.39}    & \cellcolor{gray!70}51.41          \\ 
        SAMMD-FPE & 5.60M     & \cellcolor{gray!20}81.93             & \cellcolor{gray!70}\textbf{51.45} \\ 
        SAMMD-PE & 5.61M    & \cellcolor{gray!20}82.06             & \cellcolor{gray!70}51.00         \\ 
        UNet    & 39.10M    & \cellcolor{gray!20}79.10             & \cellcolor{gray!70}49.20         \\ 
        DA      & 7.87M     & \cellcolor{gray!20}76.98             & \cellcolor{gray!70}48.51         \\ 
        ResNet  & 24.35M    & \cellcolor{gray!20}77.88             & \cellcolor{gray!70}43.60         \\ 
        SegFormer  & 4.74M     & \cellcolor{gray!20}68.55             & \cellcolor{gray!70}41.65          \\ 
        Linear  & 52.24K    & \cellcolor{gray!20}61.09             & \cellcolor{gray!70}36.10         \\ 
    \end{tabular}
\end{table}

\subsection{PEFT method selection}
We conducted this experiment to identify the optimal fine-tuning strategy for each foundational model (FM). Similar to the decoder selection experiments, we utilized datasets of healthy brain, prostate, and lumbar spine, employing the base versions of FMs (DinoV2-B, SAM-B, MedSAM, and MAE-B). In this experiment, each dataset was used as a source dataset once, with the remaining datasets serving as targets. The FM encoders were kept frozen, and only the parameters of the PEFT methods and decoders were trained. For Ladder fine-tuning, we used the small version of DinoV2 as the tunable parallel network. In the PEFT method \emph{Freeze}, only the decoder parameters were trained.

Table \ref{tab:results_peft} presents the quantitative results averaged over all source datasets for the in-domain (ID) experiments and all source-target dataset combinations for the domain generalization (DG) results. The AV column shows the average of the ID and DG results, which we used to select the best PEFT method for each FM, as both ID and DG performances are crucial. The results indicate that different PEFT methods are more effective with different FMs. Consequently, we selected Rein-Lora for DinoV2, Rein for SAM, Ladder for MedSAM, and Freeze for MAE. Additionally, we observed that Ladder fine-tuning consistently achieved the best ID Dice scores across all backbones. The more detailed results for each backbone, dataset, and PEFT method can be found in the supplementary material.

\begin{table}[h]
\centering
\caption{Comparison of different PEFT methods on different FMs. (*) \textit{Includes only brain and lumbar spine datasets.} The best results in each category (ID, DG, and AV) are highlighted by bold.}
\label{tab:results_peft}
\footnotesize
\begin{tabular}{|c|ccc||ccc||ccc||ccc|}
\hline
\multirow{1}{*}{PEFT} & \multicolumn{3}{c||}{\cellcolor{darkgray} \color{white}DinoV2-B} & \multicolumn{3}{c||}{\cellcolor{darkgray} \color{white}SAM-B} & \multicolumn{3}{c||}{\cellcolor{darkgray} \color{white}MedSAM*} & \multicolumn{3}{c|}{\cellcolor{darkgray} \color{white}MAE-B} \\ \cline{2-13}
Method & \cellcolor{gray!20}ID & \cellcolor{gray!50}DG & \cellcolor{gray!70}AV & \cellcolor{gray!20}ID & \cellcolor{gray!50}DG & \cellcolor{gray!70}AV & \cellcolor{gray!20}ID & \cellcolor{gray!50}DG & \cellcolor{gray!70}AV & \cellcolor{gray!20}ID & \cellcolor{gray!50}DG & \cellcolor{gray!70}AV \\ \hline

\cellcolor{darkgray} \color{white}Freeze & \cellcolor{gray!20}80.72 & \cellcolor{gray!50}41.77 & \cellcolor{gray!70}61.25 & \cellcolor{gray!20}83.04 & \cellcolor{gray!50}37.96 & \cellcolor{gray!70}60.50 & \cellcolor{gray!20}85.55 & \cellcolor{gray!50}27.34 & \cellcolor{gray!70}56.45 & \cellcolor{gray!20}83.58 & \cellcolor{gray!50}\textbf{42.01} & \cellcolor{gray!70}\textbf{62.80} \\

\cellcolor{darkgray!70} \color{white} Rein-Lora & \cellcolor{gray!20}82.08 & \cellcolor{gray!50}\textbf{44.44} & \cellcolor{gray!70}\textbf{63.26} & \cellcolor{gray!20}82.93 & \cellcolor{gray!50}38.45 & \cellcolor{gray!70}60.69 & \cellcolor{gray!20}85.61 & \cellcolor{gray!50}\textbf{28.85} & \cellcolor{gray!70}57.23 & \cellcolor{gray!20}82.63 & \cellcolor{gray!50}40.63 & \cellcolor{gray!70}61.63 \\

\cellcolor{darkgray} \color{white} Rein & \cellcolor{gray!20}81.83 & \cellcolor{gray!50}44.42 & \cellcolor{gray!70}63.13 & \cellcolor{gray!20}82.68 & \cellcolor{gray!50}\textbf{39.00} & \cellcolor{gray!70}\textbf{60.84} & \cellcolor{gray!20}85.64 & \cellcolor{gray!50}26.98 & \cellcolor{gray!70}56.31 & \cellcolor{gray!20}82.59 & \cellcolor{gray!50}39.06 & \cellcolor{gray!70}60.83 \\

\cellcolor{darkgray!70} \color{white} Ladder & \cellcolor{gray!20}\textbf{84.95} & \cellcolor{gray!50}33.53 & \cellcolor{gray!70}59.24 & \cellcolor{gray!20}\textbf{84.48} & \cellcolor{gray!50}32.70 & \cellcolor{gray!70}58.59 & \cellcolor{gray!20}\textbf{87.07} & \cellcolor{gray!50}28.82 & \cellcolor{gray!70}\textbf{57.95} & \cellcolor{gray!20}\textbf{83.92} & \cellcolor{gray!50}36.12 & \cellcolor{gray!70}60.02
\end{tabular}
\end{table}

\begin{table}[h]
\centering
\caption{Summary of the ID and DG results of FMs for each dataset. The best three results in each category (ID, DG) and anatomy are indicated by bold.}
\footnotesize
\begin{tabular}{|c|c|c|c||c|c||c|c||c|c|}
    \hline
    \multirow{2}{*}{PEFT} & \multirow{2}{*}{FM} & \multicolumn{2}{c||}{\cellcolor{darkgray} \color{white}Healthy Brain} & \multicolumn{2}{c||}{\cellcolor{darkgray} \color{white}Lumbar Spine} & \multicolumn{2}{c||}{\cellcolor{darkgray} \color{white}Prostate} & \multicolumn{2}{c|}{\cellcolor{darkgray} \color{white}Brain Tumor} \\ \cline{3-10}
    
    & & \cellcolor{gray!20}ID & \cellcolor{gray!70}DG & \cellcolor{gray!20}ID & \cellcolor{gray!70}DG & \cellcolor{gray!20}ID & \cellcolor{gray!70}DG & \cellcolor{gray!20}ID & \cellcolor{gray!70}DG \\ \hline
    
    \cellcolor{darkgray} \color{white} Rein-Lora & \cellcolor{darkgray} \color{white} Dino-B & \cellcolor{gray!20}85.05 & \cellcolor{gray!70}\textbf{52.37} & \cellcolor{gray!20}85.27 & \cellcolor{gray!70}34.51 & \cellcolor{gray!20}75.91 & \cellcolor{gray!70}\textbf{46.44} & \cellcolor{gray!20}86.60 & \cellcolor{gray!70}44.29 \\
    
    \cellcolor{darkgray!70} \color{white} Rein & \cellcolor{darkgray!70} \color{white} SAM-B & \cellcolor{gray!20}85.45 & \cellcolor{gray!70}49.50 & \cellcolor{gray!20}85.49 & \cellcolor{gray!70}29.21 & \cellcolor{gray!20}77.09 & \cellcolor{gray!70}38.29 & \cellcolor{gray!20}86.76 & \cellcolor{gray!70}\textbf{44.94} \\ 
    
    \cellcolor{darkgray} \color{white} Ladder &  \cellcolor{darkgray} \color{white} MedSAM & \cellcolor{gray!20}86.04 & \cellcolor{gray!70}44.22 & \cellcolor{gray!20}\textbf{88.10} & \cellcolor{gray!70}13.41 & \cellcolor{gray!20}- & \cellcolor{gray!70}- & \cellcolor{gray!20}- & \cellcolor{gray!70}- \\
    
    \cellcolor{darkgray!70} \color{white} Freeze & \cellcolor{darkgray!70} \color{white} MAE-B & \cellcolor{gray!20}85.54 & \cellcolor{gray!70}47.25 & \cellcolor{gray!20}86.79 & \cellcolor{gray!70}36.17 & \cellcolor{gray!20}\textbf{78.41} & \cellcolor{gray!70}42.63 & \cellcolor{gray!20}87.30 & \cellcolor{gray!70}44.24 \\ \hline
    \hline
    \cellcolor{darkgray} \color{white} Rein-Lora & \cellcolor{darkgray} \color{white} Dino-L & \cellcolor{gray!20}84.61 & \cellcolor{gray!70}\textbf{54.09} & \cellcolor{gray!20}85.06 & \cellcolor{gray!70}31.30 & \cellcolor{gray!20}72.95 & \cellcolor{gray!70}\textbf{42.74} & \cellcolor{gray!20}\textbf{87.13} & \cellcolor{gray!70}\textbf{50.84} \\
    
    \cellcolor{darkgray!70} \color{white} Rein & \cellcolor{darkgray!70} \color{white} SAM-L & \cellcolor{gray!20}85.50 & \cellcolor{gray!70}51.78 & \cellcolor{gray!20}86.95 & \cellcolor{gray!70}38.03 & \cellcolor{gray!20}76.07 & \cellcolor{gray!70}39.01 & \cellcolor{gray!20}86.78 & \cellcolor{gray!70}44.31 \\
    
    \cellcolor{darkgray} \color{white} Freeze & \cellcolor{darkgray} \color{white} MAE-L & \cellcolor{gray!20}85.69 & \cellcolor{gray!70}47.20 & \cellcolor{gray!20}85.97 & \cellcolor{gray!70}\textbf{48.45} & \cellcolor{gray!20}\textbf{77.82} & \cellcolor{gray!70}42.77 & \cellcolor{gray!20}\textbf{87.60} & \cellcolor{gray!70}43.71 \\ \hline
    \hline
    \cellcolor{darkgray!70} \color{white} Rein-Lora & \cellcolor{darkgray!70} \color{white} Dino-G & \cellcolor{gray!20}85.13 & \cellcolor{gray!70}\textbf{54.68} & \cellcolor{gray!20}84.14 & \cellcolor{gray!70}\textbf{40.90} & \cellcolor{gray!20}74.62 & \cellcolor{gray!70}40.37 & \cellcolor{gray!20}87.15 & \cellcolor{gray!70}41.83 \\
    
    \cellcolor{darkgray} \color{white} Rein & \cellcolor{darkgray} \color{white} SAM-H & \cellcolor{gray!20}85.36 & \cellcolor{gray!70}51.70 & \cellcolor{gray!20}86.96 & \cellcolor{gray!70}33.12 & \cellcolor{gray!20}74.78 & \cellcolor{gray!70}38.80 & \cellcolor{gray!20}87.24 & \cellcolor{gray!70}\textbf{47.26} \\
    
    \cellcolor{darkgray!70} \color{white} Freeze & \cellcolor{darkgray!70} \color{white} MAE-H & \cellcolor{gray!20}\textbf{86.14} & \cellcolor{gray!70}48.55 & \cellcolor{gray!20}85.13 & \cellcolor{gray!70}\textbf{42.15} & \cellcolor{gray!20}\textbf{79.53} & \cellcolor{gray!70}\textbf{45.95} & \cellcolor{gray!20}\textbf{87.54} & \cellcolor{gray!70}44.44 \\ \hline
    \hline
    \cellcolor{darkgray} \color{white}Full Fine Tune & \cellcolor{darkgray} \color{white}Res101+UNet & \cellcolor{gray!20}86.07 & \cellcolor{gray!70}43.99 & \cellcolor{gray!20}\textbf{87.89} & \cellcolor{gray!70}2.35 & \cellcolor{gray!20}69.06 & \cellcolor{gray!70}31.46 & \cellcolor{gray!20}85.71 & \cellcolor{gray!70}31.48 \\
    
    \cellcolor{darkgray!70} \color{white}Full Fine Tune & \cellcolor{darkgray!70} \color{white}Vanilla UNet & \cellcolor{gray!20}\textbf{88.47} & \cellcolor{gray!70}44.96 & \cellcolor{gray!20}\textbf{87.02} & \cellcolor{gray!70}1.22 & \cellcolor{gray!20}69.10 & \cellcolor{gray!70}13.41 & \cellcolor{gray!20}86.17 & \cellcolor{gray!70}37.32  \\
    
    \cellcolor{darkgray} \color{white}Full Fine Tune & \cellcolor{darkgray} \color{white}Swin UNet & \cellcolor{gray!20}\textbf{86.71} & \cellcolor{gray!70}41.68 & \cellcolor{gray!20}86.79 & \cellcolor{gray!70}5.63 & \cellcolor{gray!20}65.51 & \cellcolor{gray!70}29.90 & \cellcolor{gray!20}85.66 & \cellcolor{gray!70}40.69 \\ \hline
\end{tabular}
\label{tab:id_dg_all}
\end{table}
\subsection{ID and DG Performance Evaluation of FMs}
We evaluated both ID and DG performance of FMs with different sizes (Base, Large, and Huge)\footnote{Note that MedSAM is only available in base size and DinoV2-G refers to the giant version, the largest available version of DinoV2.}. Since the NCI prostate and the BraTS datasets were used during the training of MedSAM, we did not include these datasets in the evaluations of the MedSAM model.

We compared the performance of the foundational models (FMs) with three benchmarks: Vanilla UNet, UNet with a ResNet101 backbone pretrained on the Imagenet1K dataset \cite{deng2009imagenet} (referred to as Res101+Unet), and Swin UNet \cite{cao2022swin}. 
The summary of the Dice score results are presented in Table \ref{tab:id_dg_all} and more detailed results are presented in Tables \ref{tab:id_dg_healthy_brain}, \ref{tab:id_dg_lumbar_spine}, \ref{tab:id_dg_prostate}, \ref{tab:id_dg_brain_tumor} for each dataset. The results demonstrate that FMs consistently achieve superior DG results compared to the benchmarks. We did not observe much benefit of using larger FMs in the ID results while the large and the huge models achieve slightly better DG performance. Additionally, we observed that the benchmarks perform comparably to the FMs in most datasets in terms of ID performance, with the exception of the prostate dataset. The primary benefit of using FMs in our experiments is the improvement in DG performance rather than ID performance. This enhancement can be particularly valuable when performing unsupervised domain adaptation on the target domain or semi-supervised learning, especially using the algorithms that rely on pseudolabels \cite{hoyer2022daformer, gao2022cross, chaitanya2023local}. We leave these experiments as a future work.
\begin{table}[ht]
\centering
\caption{ID and DG performance of FMs on Healthy Brain Datasets}
\label{tab:id_dg_healthy_brain}
\tiny
\resizebox{\columnwidth}{!}{%
\begin{tabular}{|c|c|c|c|c|c|c|}
\hline
Source Dataset & PEFT & Backbone & HCP-T1w & HCP-T2w & ABIDE-C & ABIDE-S \\ \hline
\multirow{13}{*}{HCP-T1w} & Rein-Lora & DinoV2-B & 86.27 & 24.87 & 78.64 & 68.38  \\ 
                          & Rein      & SAM-B    & 86.50 & 21.06 & 79.26 & 72.16 \\ 
                          & Ladder    & MedSAM   & 86.66 & 10.10 & 77.91 & 75.81 \\
                          & Freeze    & MAE-B    & 86.65 & 17.46 & 79.28 & 74.20 \\ \cline{2-7}
                          & Rein-Lora & DinoV2-L & 85.64 & 28.27 & 76.49 & 67.41 \\ 
                          & Reins     & SAM-L    & 86.49 & 26.42 & 79.92 & 72.34 \\ 
                          & Freeze    & MAE-L    & 86.73 & 14.71 & 79.92 & 74.72 \\ \cline{2-7}
                          & Rein-Lora & Dino-G   & 86.26 & 24.41 & 77.91 & 70.91 \\ 
                          & Rein      & SAM-H    & 86.35 & 25.58 & 80.33 & 72.72 \\ 
                          & Freeze    & MAE-H    & 86.97 & 14.63 & 79.72 & 75.67 \\ \cline{2-7}
                          & Full Fine Tune & Res101+UNet  & 86.92 & 8.93 & 80.79 & 75.88 \\ 
                          & N/A            & Vanilla UNet & 88.94 & 10.85 & 80.20 & 73.77 \\ 
                          & N/A            & Swin UNet    & 87.86 & 8.92 & 79.38 & 65.94 \\ \hline

\multirow{13}{*}{HCP-T2w} & Rein-Lora & DinoV2-B & 34.89 & 84.91 & 37.76 & 26.75 \\ 
                          & Rein      & SAM-B    & 23.83 & 85.34 & 24.92 & 13.56 \\ 
                          & Ladder    & MedSAM   & 10.67 & 85.66 & 10.83 & 9.13 \\
                          & Freeze    & MAE-B    & 17.01 & 85.32 & 19.18 & 12.62 \\ \cline{2-7}
                          & Rein-Lora & DinoV2-L & 41.19 & 84.19 & 40.43 & 28.80 \\ 
                          & Rein      & SAM-L    & 27.62 & 85.44 & 27.32 & 16.20 \\ 
                          & Freeze    & MAE-L    & 18.40 & 85.46 & 21.77 & 13.09 \\ \cline{2-7}
                          & Rein-Lora & DinoV2-G & 42.08 & 85.03 & 41.16 & 27.58 \\ 
                          & Rein      & SAM-H    & 28.07 & 85.41 & 29.86 & 14.62 \\ 
                          & Freeze    & MAE-H    & 20.31 & 85.95 & 19.26 & 14.16 \\ \cline{2-7}
                          & Full Fine Tune & Res101+UNet  & 11.25 & 86.34 & 8.20 & 7.75  \\ 
                          & N/A            & Vanilla UNet & 12.02 & 87.92 & 11.73 & 8.54 \\ 
                          & N/A            & Swin UNet    & 11.57 & 86.68 & 14.92 & 9.17 \\ \hline

\multirow{13}{*}{ABIDE-C} & Rein-Lora & DinoV2-B & 78.15 & 31.02 & 85.79 & 72.85 \\ 
                         & Rein      & SAM-B    & 79.64 & 36.32 & 86.13 & 74.73 \\ 
                         & Ladder    & MedSAM   & 79.03 & 10.62 & 86.98 & 77.41 \\
                         & Freeze    & MAE-B    & 79.37 & 24.71 & 86.20 & 71.35 \\ \cline{2-7}
                         & Rein-Lora & DinoV2-L & 78.46 & 34.84 & 85.50 & 74.35 \\ 
                         & Rein      & SAM-L    & 79.54 & 37.57 & 86.20 & 75.06 \\ 
                         & Freeze    & MAE-L    & 79.24 & 27.03 & 86.36 & 74.51 \\ \cline{2-7}
                         & Rein-Lora & DinoV2-G & 78.98 & 37.68 & 86.12 & 75.84 \\ 
                         & Rein      & SAM-H    & 79.41 & 38.29 & 86.22 & 76.97 \\ 
                         & Freeze    & MAE-H    & 79.54 & 29.86 & 87.01 & 77.76 \\ \cline{2-7}
                         & Full Fine Tune & Res101+UNet  & 77.91 & 7.36 & 87.13 & 78.76 \\ 
                         & N/A            & Vanilla UNet & 78.58 & 11.37 & 89.58 & 79.96 \\ 
                         & N/A            & Swin UNet & 77.83 & 15.65 & 87.25 & 72.22 \\ \hline

\multirow{13}{*}{ABIDE-S} & Rein-Lora & DinoV2-B & 72.68 & 21.63 & 80.87 & 83.23 \\ 
                          & Rein      & SAM-B    & 74.81 & 12.26 & 81.49 & 83.84 \\ 
                          & Ladder    & MedSAM-B & 77.60 & 9.46 &	82.07 &	84.87 \\
                          & Freeze    & MAE-B    & 74.83 & 15.11 & 81.82 & 84.00 \\ \cline{2-7}
                          & Rein-Lora & DinoV2-L & 73.99 & 25.78 & 79.11 & 83.12 \\ 
                          & Rein      & SAM-L    & 75.27 & 22.47 & 81.68 & 83.86 \\ 
                          & Freeze    & MAE-L    & 71.44 & 10.30 & 81.25 & 84.20 \\ \cline{2-7}
                          & Rein-Lora & DinoV2-G & 73.86 & 26.06 & 79.72 & 83.09 \\ 
                          & Rein      & SAM-H    & 73.23 & 19.63 & 81.67 & 83.47 \\ 
                          & Freeze    & MAE-H    & 73.90 & 15.10 & 82.66 & 84.61 \\  \cline{2-7}
                          & Full Fine Tune  & Res101+UNet & 79.02 & 9.35 & 82.70 & 83.89 \\ 
                          & N/A             & Vanilla UNet & 78.69 & 8.32 & 85.52 & 87.42 \\ 
                          & N/A             & Swin UNet & 64.29 & 6.59 & 73.70 & 85.03 \\ \hline
\end{tabular}
}
\end{table}
\begin{table}[h]
\centering
\caption{ID and DG performance of FMs on Lumbar Spine Datasets}
\label{tab:id_dg_lumbar_spine}
\tiny
\begin{tabular}{|c|c|c|c|c|}
\hline
Source Dataset & PEFT & Backbone & VerSe & MRSegV \\ \hline
\multirow{13}{*}{VerSe} & Rein-Lora & DinoV2-B & 83.87 & 36.09 \\ 
                        & Rein      & SAM-B    & 82.99 & 23.10 \\ 
                        & Ladder    & MedSAM-B & 88.17 & 2.54 \\
                        & Freeze    & MAE-B    & 86.33 & 32.54 \\ \cline{2-5}
                        & Rein-Lora & DinoV2-L & 83.61 & 15.82 \\ 
                        & Rein      & SAM-L    & 86.24 & 35.94 \\ 
                        & Freeze    & MAE-L    & 83.99 & 53.58 \\ \cline{2-5}
                        & Rein-Lora & DinoV2-G & 81.28 & 41.20 \\ 
                        & Rein      & SAM-H    & 86.53 & 33.63 \\ 
                        & Freeze    & MAE-H    & 82.12 & 43.49 \\ \cline{2-5}
                        & Full Fine Tune & Res101+UNet & 87.82 & 0.00 \\ 
                        & N/A            & Vanilla UNet & 85.97 & 0.21 \\ 
                        & N/A            & Swin UNet & 86.63 & 1.22 \\ \hline
\multirow{13}{*}{MRSegV}  & Rein-Lora & DinoV2-B & 32.92 & 86.67 \\ 
                          & Rein      & SAM-B    & 35.31 & 87.99 \\ 
                          & Ladder    & MedSAM-B & 24.28 & 88.02 \\
                          & Freeze    & MAE-B    & 39.79 & 87.25 \\ \cline{2-5}
                          & Rein-Lora & DinoV2-L & 46.77 & 86.51 \\ 
                          & Rein      & SAM-L    & 40.12 & 87.66 \\ 
                          & Freeze    & MAE-L    & 43.31 & 87.94 \\ \cline{2-5}
                          & Rein-Lora & DinoV2-G & 40.59 & 87.00 \\ 
                          & Rein      & SAM-H    & 32.60 & 87.39 \\ 
                          & Freeze    & MAE-H    & 40.80 & 88.13 \\ \cline{2-5}
                          & Full Fine Tune & Res101+UNet & 4.70 & 87.96 \\ 
                          & N/A & Vanilla UNet & 2.22 & 88.07 \\ 
                          & N/A & Swin UNet & 10.03 & 86.95 \\ \hline
\end{tabular}
\end{table}
\begin{table}[h]
\centering
\caption{ID and DG performance of FMs on Prostate Datasets}
\label{tab:id_dg_prostate}
\tiny
\begin{tabular}{|c|c|c|c|c|}
\hline
Source Dataset & PEFT & Backbone & NCI & In-house \\ \hline
\multirow{12}{*}{NCI} & Rein-Lora & DinoV2-B & 69.19 & 31.11 \\ 
                      & Rein      & SAM-B    & 68.77 & 23.49 \\ 
                      & Freeze    & MAE-B    & 72.98 & 29.28 \\ \cline{2-5}
                      & Rein-Lora & DinoV2-L & 68.72 & 28.04 \\ 
                      & Rein      & SAM-L    & 67.58 & 25.13 \\ 
                      & Freeze    & MAE-L    & 72.07 & 26.20 \\ \cline{2-5}
                      & Rein-Lora & DinoV2-G & 70.63 & 27.93 \\ 
                      & Rein      & SAM-H    & 65.62 & 23.41 \\ 
                      & Freeze    & MAE-H    & 73.85 & 30.00 \\ \cline{2-5}
                      & Full Fine Tune & Res101+UNet & 63.46 & 18.98 \\ 
                      & N/A            & Vanilla UNet & 75.33 & 25.39 \\ 
                      & N/A            & Swin UNet & 58.66 & 25.09 \\ \hline
\multirow{12}{*}{In-house} & Rein-Lora & DinoV2-B & 61.76 & 82.62 \\ 
                           & Rein      & SAM-B    & 53.08 & 85.40 \\ 
                           & Freeze    & MAE-B    & 55.98 & 83.84 \\ \cline{2-5}
                           & Rein-Lora & DinoV2-L & 57.43 & 77.18 \\ 
                           & Rein      & SAM-L    & 52.88 & 84.55 \\ 
                           & Freeze    & MAE-L    & 59.33 & 83.57 \\ \cline{2-5}
                           & Rein-Lora & DinoV2-G & 52.80 & 78.60 \\ 
                           & Rein      & SAM-H    & 54.18 & 83.94 \\ 
                           & Freeze    & MAE-H    & 61.90 & 85.21 \\ \cline{2-5}
                           & Full Fine Tune & Res101+UNet & 43.94 & 74.66 \\ 
                           & N/A & Vanilla UNet & 1.43 & 62.87 \\ 
                           & N/A & Swin UNet & 34.70 & 72.35 \\ 
\hline
\end{tabular}
\end{table}
\begin{table}[h]
\centering
\caption{ID and DG performance of FMs on Brain Tumor Datasets}
\label{tab:id_dg_brain_tumor}
\tiny
\begin{tabular}{|c|c|c|c|c|}
\hline
Source Dataset & PEFT & Backbone & BraTS-T1w & BraTS-FLAIR \\ \hline
\multirow{12}{*}{BraTS-T1w} & Rein-Lora & DinoV2-B & 82.14 & 60.96 \\ 
                            & Rein      & SAM-B    & 82.34 & 55.06 \\ 
                            & Freeze    & MAE-B    & 83.07 & 60.23 \\ \cline{2-5}
                            & Rein-Lora & DinoV2-L & 82.95 & 63.49 \\ 
                            & Rein      & SAM-L    & 82.26 & 57.39 \\ 
                            & Freeze    & MAE-L    & 83.30 & 57.30 \\ \cline{2-5}
                            & Rein-Lora & DinoV2-G & 82.73 & 60.28 \\ 
                            & Rein      & SAM-H    & 82.95 & 58.60 \\ 
                            & Freeze    & MAE-H    & 83.37 & 60.89 \\ \cline{2-5}
                            & Full Fine Tune & Res101+UNet & 81.22 & 53.82 \\ 
                            & N/A            & Vanilla UNet & 81.86 & 54.84 \\ 
                            & N/A            & Swin UNet & 79.43 & 52.25 \\ \hline
\multirow{12}{*}{BraTS-FLAIR} & Rein-Lora & DinoV2-V & 27.62 & 91.05 \\ 
                              & Rein      & SAM-B    & 34.82 & 91.18 \\ 
                              & Freeze    & MAE-B    & 28.24 & 91.53 \\ \cline{2-5}
                              & Rein-Lora & DinoV2-L & 38.19 & 91.30 \\ 
                              & Rein      & SAM-L    & 31.23 & 91.30 \\ 
                              & Freeze    & MAE-L    & 30.11 & 91.90 \\ \cline{2-5}
                              & Rein-Lora & DinoV2-G & 23.37 & 91.57 \\ 
                              & Rein      & SAM-H    & 35.91 & 91.52 \\ 
                              & Freeze    & MAE-H    & 27.98 & 91.70 \\ \cline{2-5}
                              & Full Fine Tune & Res101+UNet & 9.14 & 90.20 \\ 
                              & N/A            & Vanilla UNet & 19.79 & 90.48 \\ 
                              & N/A            & Swin UNet & 29.12 & 91.89 \\
                              \hline
\end{tabular}
\end{table}

\section{Conclusion}
In this study, we explored the domain generalization performance of state-of-the-art FMs, DinoV2, SAM, MedSAM, and MAE, in medical image segmentation. We fine-tuned these models using various PEFT techniques, including Ladder, Rein, and Rein-LoRA, and employed different decoder heads. Our primary focus was on improving domain generalization, a critical challenge in medical imaging due to the variability in acquisition settings across different scanner models and protocols.

We proposed a decoder head named HQHSAM, which simply integrates elements from the HSAM and HQSAM architectures, demonstrating slightly enhanced domain generalization (DG) performance. Through extensive experiments on multiple datasets covering various anatomies and modalities, we showed that foundational models (FMs) significantly improve DG performance for medical image segmentation compared to benchmarks, while maintaining comparable in-domain (ID) performance. Our experiments indicate that the primary advantage of foundational models lies in enhancing DG accuracy in medical image segmentation. It is worth investigating the benefits of this improved DG performance in tasks such as unsupervised domain adaptation and semi-supervised learning, especially when using algorithms that utilize pseudolabels.

Another key finding of our study is that the effectiveness of PEFT techniques varies across different foundational models (FMs). This underscores the importance of selecting the appropriate fine-tuning strategy to optimize performance in diverse clinical settings.

In conclusion, our work emphasizes the potential of foundational models (FMs) to improve the robustness and generalization capabilities of neural networks in medical image segmentation. These findings lay a solid foundation for future research, paving the way for further advancements in training networks with enhanced generalization accuracy for medical image segmentation.

\section{Acknowledgement}
This study was financially supported by: 1. The LOOP Zürich – Medical Research Center, Zurich,
Switzerland, 2. Personalized Health and Related Technologies (PHRT), project number 222 and 643, ETH
domain, 3. ETH AI Center, and 4. Swiss National Science Foundation (SNF), project number 205320\_200877. 


%
%
\bibliographystyle{splncs04}
\bibliography{main}

\clearpage
\newpage

\appendix

\section{Detailed experimental results}

\begin{table}[h]
\centering
\caption{Detailed decoder selection experiments}
\centering
\resizebox{\textwidth}{!}{
\begin{tabular}{|c|c|c|c|c|c|c|c|c|c|}
\hline
Decoder & Encoder & MRSegV & VerSe & ABIDE-C & ABIDE-S & HCP-T1w & HCP-T2w & In-house & NCI \\ \hline

\multirow{4}{*}{Linear} & DinoV2 & 76.38 & 30.46 & 70.88 & 59.87 & 62.29 & 36.90 & 40.40 & 15.70 \\ 
                        & SAM    & 77.10 & 21.72 & 69.62 & 50.48 & 54.04 & 39.42 & 41.83 & 34.89 \\ 
                        & MedSAM & 68.50 & 5.24 & 68.02 & 35.74 & 50.45 & 11.85 & 29.98 & -- \\ 
                        & MAE    & 75.38 & 24.86 & 66.85 & 49.77 & 51.45 & 27.26 & 48.17 & 33.70  \\ \hline

\multirow{4}{*}{SegFormer} & DinoV2 & 82.80 & 38.50 & 75.63 & 64.51 & 68.22 & 37.31 & 45.32 & 14.59 \\ 
                           & SAM    & 83.71 & 28.57 & 76.16 & 63.33 & 62.84 & 33.99 & 52.98 & 29.85 \\ 
                           & MedSAM & 80.27 & 12.51 & 75.55 & 56.48 & 63.44 & 14.31 & 41.00 & -- \\ 
                           & MAE    & 82.34 & 40.47 & 73.71 & 60.78 & 65.42 & 21.06 & 53.16 & 20.05 \\ \hline

\multirow{4}{*}{DA} & DinoV2 & 85.05 & 45.02 & 78.77 & 70.57 & 72.58 & 46.44 & 64.95 & 45.21 \\ 
                    & SAM    & 84.63 & 32.79 & 77.98 & 69.47 & 68.93 & 44.06 & 72.96 & 23.80 \\ 
                    & MedSAM & 82.24 & 13.97 & 77.00 & 64.02 & 67.50 & 15.23 & 59.17 & -- \\ 
                    & MAE    & 83.51 & 38.56 & 75.37 & 65.70 & 68.40 & 29.14 & 82.16 & 48.71 \\ \hline

\multirow{4}{*}{SAMMD-FPE} & DinoV2 & 84.64 & 44.58 & 78.12 & 60.26 & 72.04 & 46.48 & 77.88 & 45.69 \\
                           & SAM    & 86.99 & 31.07 & 85.05 & 58.13 & 77.03 & 62.85 & 76.38 & 47.81 \\
                           & MedSAM & 84.65 & 37.47 & 85.23 & 49.10 & 73.75 & 18.02 & 72.97 & -- \\
                           & MAE    & 87.10 & 35.72 & 84.19 & 68.94 & 77.99 & 17.40 & 81.52 & 51.03 \\ \hline

\multirow{4}{*}{SAMMD-PE} & DinoV2 & 84.30 & 45.78 & 78.31 & 60.64 & 72.06 & 48.07 & 78.10 & 52.10 \\
                          & SAM    & 87.22 & 26.84 & 85.19 & 59.03 & 77.23 & 64.01 & 75.07 & 38.65 \\
                          & MedSAM & 84.65 & 34.89 & 85.17 & 44.68 & 74.79 & 21.81 & 75.94 & -- \\
                          & MAE    & 87.14 & 38.34 & 84.50 & 69.29 & 77.74 & 21.46 & 77.51 & 57.51 \\ \hline

\multirow{4}{*}{HQSAM} & DinoV2 & 86.02 & 36.12 & 84.68 & 69.30 & 77.00 & 27.37 & 76.74 & 56.52 \\
                       & SAM    & 87.58 & 34.68 & 85.66 & 73.86 & 78.59 & 29.79 & 85.73 & 53.17 \\
                       & MedSAM & 86.26 & 27.86 & 85.02 & 65.83 & 77.89 & 16.16 & 83.40 & -- \\
                       & MAE    & 87.52 & 35.34 & 85.02 & 72.10 & 77.77 & 25.56 & 84.91 & 61.72 \\ \hline

\multirow{4}{*}{HSAM} & DinoV2 & 83.84 & 42.30 & 77.06 & 58.48 & 71.01 & 42.40 & 77.12 & 58.80 \\
                      & SAM    & 87.19 & 38.09 & 84.64 & 70.60 & 78.00 & 30.17 & 85.06 & 50.45 \\
                      & MedSAM & 85.53 & 29.84 & 83.94 & 55.34 & 76.99 & 15.19 & 82.88 & -- \\
                      & MAE    & 87.20 & 37.01 & 84.53 & 69.63 & 77.80 & 25.01 & 81.72 & 56.84 \\ \hline

\multirow{4}{*}{HQHSAM} & DinoV2 & 86.51 & 37.42 & 85.06 & 70.42 & 77.64 & 29.99 & 78.21 & 56.66 \\
                        & SAM    & 87.50 & 36.20 & 86.20 & 75.23 & 79.18 & 34.65 & 84.67 & 51.34 \\ 
                        & MedSAM & 86.55 & 26.68 & 85.54 & 63.54 & 77.38 & 18.61 & 82.25 & -- \\ 
                        & MAE    & 87.25 & 39.79 & 86.20 & 71.35 & 79.37 & 24.71 & 83.84 & 55.98 \\ \hline

\multirow{4}{*}{ResNet} & DinoV2 & 87.35 & 25.17 & 85.76 & 78.64 & 78.47 & 32.83 & 57.14 & 11.00 \\ 
                        & SAM    & 86.96 & 28.53 & 87.42 & 79.21 & 76.31 & 20.52 & 65.22 & 23.23 \\ 
                        & MedSAM & 85.68 & 3.12 & 87.21 & 77.68 & 77.29 & 11.08 & 49.72 & -- \\ 
                        & MAE    & 87.68 & 12.55 & 88.40 & 80.38 & 80.38 & 10.21 & 65.98 & 23.18 \\ \hline

\multirow{4}{*}{UNet} & DinoV2 & 88.46 & 22.18 & 88.62 & 81.71 & 79.88 & 31.64 & 58.65 & 42.31 \\ 
                      & SAM    & 87.37 & 26.58 & 87.75 & 78.93 & 76.38 & 30.52 & 73.44 & 40.32 \\ 
                      & MedSAM & 86.27 & 3.04 & 87.63 & 77.93 & 76.77 & 12.85 & 56.15 & -- \\ 
                      & MAE    & 87.98 & 29.40 & 88.62 & 81.99 & 80.20 & 14.99 & 58.28 & 53.76 \\ \hline
\end{tabular}
}
\end{table}

\begin{table}[h]
\centering
\caption{Detailed PEFT method selection experiments on Healthy Brain datasets}
\scriptsize
\begin{tabular}{|c|c|c|c|c|c|c|}
\hline
PEFT Method & Encoder & Source dataset & HCP-T1w & HCP-T2w & ABIDE-C & ABIDE-S \\ \hline

\multirow{16}{*}{Freeze} & \multirow{4}{*}{DinoV2} & HCP-T1w & 85.01 & 19.16 & 76.75 & 64.24 \\ 
                        &                          & HCP-T2w & 34.22 & 84.45 & 37.23 & 23.45 \\ 
                        &                          & ABIDE-C & 77.64 & 29.99 & 85.06 & 70.42 \\ 
                        &                          & ABIDE-S & 68.68 & 15.82 & 76.96 & 82.16 \\ \cline{2-7}
                        & \multirow{4}{*}{SAM}     & HCP-T1w & 86.42 & 22.71 & 79.27 & 73.09 \\ 
                        &                          & HCP-T2w & 23.35 & 85.29 & 23.95 & 14.59 \\ 
                        &                          & ABIDE-C & 79.18 & 34.65 & 86.20 & 75.23 \\ 
                        &                          & ABIDE-S & 75.64 & 15.11 & 81.82 & 83.80 \\ \cline{2-7}
                        & \multirow{4}{*}{MedSAM}  & HCP-T1w & 86.00 & 8.98 & 77.53 & 64.90  \\ 
                        &                          & HCP-T2w & 11.15 & 84.58 & 14.59 & 9.26 \\ 
                        &                          & ABIDE-C & 77.38 & 18.61 & 85.54 & 63.54 \\ 
                        &                          & ABIDE-S & 64.89 & 6.65 & 72.00 & 82.93 \\ \cline{2-7}
                        & \multirow{4}{*}{MAE}     & HCP-T1w & 86.65 & 17.46 & 79.28 & 74.20 \\ 
                        &                          & HCP-T2w & 17.01 & 85.32 & 19.18 & 12.62 \\ 
                        &                          & ABIDE-C & 79.37 & 24.71 & 86.20 & 71.35 \\ 
                        &                          & ABIDE-S & 74.83 & 15.11 & 81.82 & 84.00 \\ \hline
\multirow{16}{*}{Rein-Lora} & \multirow{4}{*}{DinoV2} & HCP-T1w & 86.27 & 24.87 & 78.64 & 68.38 \\ 
                            &                         & HCP-T2w & 34.89 & 84.91 & 37.76 & 26.75 \\ 
                            &                         & ABIDE-C & 78.15 & 31.02 & 85.79 & 72.85 \\ 
                            &                         & ABIDE-S & 72.68 & 21.63 & 80.87 & 83.23 \\ \cline{2-7}
                            & \multirow{4}{*}{SAM}    & HCP-T1w & 86.53 & 22.52 & 79.13 & 74.51 \\ 
                            &                         & HCP-T2w & 21.75 & 85.38 & 20.74 & 12.86 \\ 
                            &                         & ABIDE-C & 79.64 & 37.42 & 86.11 & 74.78 \\ 
                            &                         & ABIDE-S & 73.02 & 14.48 & 82.12 & 83.64 \\ \cline{2-7}
                            & \multirow{4}{*}{MedSAM} & HCP-T1w & 86.31 & 10.60 & 79.23 & 69.88 \\ 
                            &                         & HCP-T2w & 11.33 & 84.53 & 11.54 & 9.13 \\ 
                            &                         & ABIDE-C & 77.72 & 19.84 & 85.63 & 68.07 \\ 
                            &                         & ABIDE-S & 69.96 & 6.70 & 75.81 & 83.40 \\ \cline{2-7}
                            & \multirow{4}{*}{MAE}    & HCP-T1w & 86.65 & 16.10 & 79.46 & 74.18 \\ 
                            &                         & HCP-T2w & 19.25 & 85.41 & 22.00 & 13.09 \\ 
                            &                         & ABIDE-C & 78.94 & 22.94 & 86.25 & 74.32 \\ 
                            &                         & ABIDE-S & 73.72 & 17.08 & 80.64 & 83.57 \\ \hline
\multirow{16}{*}{Rein} & \multirow{4}{*}{DinoV2} & HCP-T1w & 86.34 & 25.15 & 78.59 & 70.87 \\ 
                       &                         & HCP-T2w & 35.30 & 84.83 & 38.45 & 23.79 \\ 
                       &                         & ABIDE-C & 77.90 & 31.68 & 85.91 & 74.32 \\ 
                       &                         & ABIDE-S & 72.66 & 20.32 & 80.44 & 83.14 \\ \cline{2-7}
                       & \multirow{4}{*}{SAM} & HCP-T1w & 86.50 & 21.06 & 79.26 & 72.16 \\ 
                       &                      & HCP-T2w & 23.83 & 85.34 & 24.92 & 13.56 \\ 
                       &                      & ABIDE-C & 79.64 & 36.32 & 86.13 & 74.73 \\ 
                       &                      & ABIDE-S & 74.81 & 12.26 & 81.49 & 83.84 \\ \cline{2-7}
                       & \multirow{4}{*}{MedSAM} & HCP-T1w & 86.38 & 8.69 & 77.70 & 63.37 \\ 
                       &                         & HCP-T2w & 12.56 & 84.52 & 12.94 & 9.64 \\ 
                       &                         & ABIDE-C & 77.54 & 15.59 & 85.44 & 65.06 \\ 
                       &                         & ABIDE-S & 67.08 & 6.62 & 75.64 & 83.43 \\ \cline{2-7}
                       & \multirow{4}{*}{MAE}    & HCP-T1w & 86.30 & 17.20 & 80.24 & 74.98 \\ 
                       &                         & HCP-T2w & 15.13 & 85.42 & 17.45 & 12.17 \\ 
                       &                         & ABIDE-C & 79.21 & 26.51 & 86.23 & 74.57 \\ 
                       &                         & ABIDE-S & 74.09 & 13.70 & 80.52 & 83.66 \\ \hline
\multirow{16}{*}{Ladder} & \multirow{4}{*}{DinoV2} & HCP-T1w & 87.11 & 10.52 & 79.21 & 76.62 \\ 
                         &                         & HCP-T2w & 11.50 & 86.37 & 13.59 & 9.54 \\ 
                         &                         & ABIDE-C & 78.42 & 16.75 & 87.54 & 78.33 \\ 
                         &                         & ABIDE-S & 79.55 & 10.85 & 83.14 & 85.45 \\ \cline{2-7}
                         & \multirow{4}{*}{SAM}    & HCP-T1w & 86.74 & 11.03 & 78.80 & 75.76 \\ 
                         &                         & HCP-T2w & 9.28 & 85.85 & 9.20 & 7.76 \\ 
                         &                         & ABIDE-C & 78.95 & 13.14 & 86.98 & 78.17 \\ 
                         &                         & ABIDE-S & 77.31 & 12.01 & 81.92 & 84.74 \\ \cline{2-7}
                         & \multirow{4}{*}{MedSAM} & HCP-T1w & 86.66 & 10.10 & 77.91 & 75.81 \\ 
                         &                         & HCP-T2w & 10.67 & 85.66 & 10.83 & 9.13 \\ 
                         &                         & ABIDE-C & 79.03 & 10.62 & 86.98 & 77.41 \\ 
                         &                         & ABIDE-S & 77.60 & 9.46 & 82.07 & 84.87 \\ \cline{2-7}
                         & \multirow{4}{*}{MAE}    & HCP-T1w & 86.54 & 11.01 & 79.09 & 75.67 \\ 
                         &                         & HCP-T2w & 11.53 & 85.71 & 15.48 & 11.14 \\ 
                         &                         & ABIDE-C & 78.65 & 17.83 & 86.73 & 77.83 \\ 
                         &                         & ABIDE-S & 77.00 & 9.66 & 82.48 & 84.42 \\ 
\hline
\end{tabular}
\end{table}

\begin{table}[h]
\centering
\caption{Detailed PEFT method selection experiments on Lumbar Spine datasets}
\footnotesize

\begin{tabular}{|c|c|c|c|c|}
\hline
PEFT Method & Encoder & Source dataset & VerSe & MRSegV \\ \hline

\multirow{8}{*}{Freeze} & \multirow{2}{*}{DinoV2} & VerSe  & 83.77 & 30.08 \\ 
                        &                         & MRSegV & 37.42 & 86.51 \\ \cline{2-5}
                        & \multirow{2}{*}{SAM}    & VerSe  & 86.45 & 18.10 \\ 
                        &                         & MRSegV & 36.20 & 87.50 \\ \cline{2-5}
                        & \multirow{2}{*}{MedSAM} & VerSe  & 86.14 & 1.09  \\ 
                        &                         & MRSegV & 26.68 & 86.55 \\ \cline{2-5}
                        & \multirow{2}{*}{MAE}    & VerSe  & 86.33 & 32.54 \\ 
                        &                         & MRSegV & 39.79 & 87.25 \\ \hline

\multirow{8}{*}{Rein-Lora} & \multirow{2}{*}{DinoV2} & VerSe  & 83.87 & 36.09 \\ 
                           &                         & MRSegV & 32.92 & 86.67 \\ \cline{2-5}
                           & \multirow{2}{*}{SAM}    & VerSe  & 86.60 & 19.04 \\ 
                           &                         & MRSegV & 36.20 & 87.80 \\ \cline{2-5}
                           & \multirow{2}{*}{MedSAM} & VerSe  & 86.35 & 1.43  \\ 
                           &                         & MRSegV & 29.02 & 86.16 \\ \cline{2-5}
                           & \multirow{2}{*}{MAE}    & VerSe  & 86.27 & 25.94 \\ 
                           &                         & MRSegV & 39.56 & 87.61 \\ \hline

\multirow{8}{*}{Rein}      & \multirow{2}{*}{DinoV2} & VerSe  & 84.14 & 32.66 \\ 
                           &                         & MRSegV & 37.01 & 86.27 \\ \cline{2-5}
                           & \multirow{2}{*}{SAM}    & VerSe  & 82.99 & 23.10 \\ 
                           &                         & MRSegV & 35.31 & 87.99 \\ \cline{2-5}
                           & \multirow{2}{*}{MedSAM} & VerSe  & 86.30 & 2.38 \\ 
                           &                         & MRSegV & 23.45 & 86.38 \\ \cline{2-5}
                           & \multirow{2}{*}{MAE}    & VerSe  & 83.54 & 15.70 \\ 
                           &                         & MRSegV & 38.28 & 87.55 \\ \hline

\multirow{8}{*}{Ladder} & \multirow{2}{*}{DinoV2}    & VerSe  & 87.82 & 3.11 \\ 
                        &                            & MRSegV & 22.19 & 87.73 \\ \cline{2-5}
                        & \multirow{2}{*}{SAM}       & VerSe  & 88.20 & 12.22 \\ 
                        &                            & MRSegV & 9.87 & 87.91 \\ \cline{2-5}
                        & \multirow{2}{*}{MedSAM}    & VerSe  & 88.17 & 2.54 \\ 
                        &                            & MRSegV & 24.28 & 88.02 \\ \cline{2-5}
                        & \multirow{2}{*}{MAE}       & VerSe  & 86.75 & 11.03 \\ 
                        &                            & MRSegV & 26.79 & 87.61 \\ 
\hline
\end{tabular}
\end{table}

\begin{table}[h]
\centering
\caption{Detailed PEFT method selection experiments on Prostate datasets}
\footnotesize
\begin{tabular}{|c|c|c|c|c|}
\hline
PEFT Method & Encoder & Source dataset & NCI & In-house \\ \hline

\multirow{6}{*}{Freeze} & \multirow{2}{*}{DinoV2} & NCI      & 67.51 & 27.37 \\ 
                        &                         & In-house & 56.66 & 78.21 \\ \cline{2-5}
                        & \multirow{2}{*}{SAM}    & NCI      & 68.78 & 22.36 \\ 
                        &                         & In-house & 51.34 & 84.67 \\ \cline{2-5}
                        & \multirow{2}{*}{MAE}    & NCI      & 72.98 & 29.28 \\ 
                        &                         & In-house & 55.98 & 83.84 \\ \hline

\multirow{6}{*}{Rein-Lora} & \multirow{2}{*}{DinoV2} & NCI      & 69.19 & 31.11 \\ 
                           &                         & In-house & 61.76 & 82.62 \\ \cline{2-5}
                           & \multirow{2}{*}{SAM}    & NCI      & 66.63 & 21.89 \\ 
                           &                         & In-house & 54.77 & 85.71 \\ \cline{2-5}
                           & \multirow{2}{*}{MAE}    & NCI      & 68.95 & 27.95 \\ 
                           &                         & In-house & 55.02 & 82.01 \\ \hline

\multirow{6}{*}{Rein}      & \multirow{2}{*}{DinoV2} & NCI      & 68.71 & 31.01 \\ 
                           &                         & In-house & 60.93 & 81.76 \\ \cline{2-5}
                           & \multirow{2}{*}{SAM}    & NCI      & 68.77 & 23.49 \\ 
                           &                         & In-house & 53.08 & 85.40 \\ \cline{2-5}
                           & \multirow{2}{*}{MAE}    & NCI      & 68.71 & 27.40 \\ 
                           &                         & In-house & 58.68 & 84.93 \\ \hline

\multirow{6}{*}{Ladder} & \multirow{2}{*}{DinoV2}    & NCI      & 76.86 & 25.29 \\ 
                        &                            & In-house & 59.23 & 84.03 \\ \cline{2-5}
                        & \multirow{2}{*}{SAM}       & NCI      & 72.80 & 24.78 \\ 
                        &                            & In-house & 60.44 & 85.80 \\ \cline{2-5}
                        & \multirow{2}{*}{MAE}       & NCI      & 72.04 & 26.21 \\ 
                        &                            & USZ      & 61.49 & 85.43 \\ 
\hline
\end{tabular}
\label{tab:summary_fine_tuning}
\end{table}

\end{document}